# H1B-KV: Hybrid One-Bit Caches for Memory-Efficient Large Language Model Inference

Harshil Vejendla
Department of Computer Science
Rutgers University
New Brunswick, NJ, USA
harshil.vejendla@rutgers.edu

*Abstract*—Autoregressive decoding in large language models (LLMs) requires caching a growing list of past key-value (KV) pairs, making long-context inference a memory-bound problem. While recent methods have explored quantizing the cache, evicting tokens, or using binary sketches for keys (e.g., Loki), these approaches often provide an incomplete solution by leaving one component (like values) uncompressed or by discarding context information. This paper introduces the Hybrid One-Bit KV Cache (H1B-KV), a comprehensive compression scheme that radically reduces memory usage without sacrificing context. H1B-KV represents each key vector using a 1-bit binary sketch, enabling hardware-friendly bitwise attention, and further compresses value vectors using 4-bit quantization. This holistic, hybrid approach allows a 7-billion parameter LLM to handle an 8k-token context with under 60 MB of cache memory—a 70x reduction. We demonstrate that after a lightweight finetuning, H1B-KV matches full-precision performance not only on perplexity benchmarks but also on complex downstream tasks like mathematical reasoning (GSM8K), multi-task understanding (MMLU), and code generation (HumanEval). Our results show H1B-KV significantly outperforms leading quantization (KIVI), token eviction (SparseLLM), and key-only sketching (Loki) methods in quality-per-byte, establishing it as a robust solution for deploying LLMs in memory-constrained environments.

## I. INTRODUCTION

The success of large language models (LLMs) like GPT and Llama is driven by the Transformer architecture [1]. During autoregressive generation, these models must store the key (K) and value (V) vectors for every token in the context history. This KV cache grows linearly with the sequence length, consuming enormous amounts of memory. For a 7B parameter model, the cache for a 32k-token context can exceed 16 GB in standard 16-bit precision (FP16), rendering deployment on memory-constrained edge devices like smartphones and embedded systems infeasible.

Existing cache-compression techniques aim to mitigate this memory bottleneck. Methods include quantizing KV pairs to lower bit-widths (e.g., 4-bit) [2], [3], factorizing the cache into low-rank matrices [4], or selectively evicting less important tokens [5]. While effective, these approaches still often store multi-byte floating-point or integer representations, leaving a considerable memory and bandwidth footprint.

This work poses a more aggressive question: can LLMs operate effectively if we replace each high-dimensional key vector with an extreme binary sketch, and further compress the corresponding value vector? We answer in the affirmative by proposing the **Hybrid One-Bit KV Cache (H1B-KV)**. Our main contributions are:

1) We introduce H1B-KV, a novel cache architecture where keys are compressed to a 1-bit random projection sketch and values are compressed using 4-bit quantization. This enables a dramatic 98% reduction in cache size.
2) We provide a theoretical justification for our method, linking the binary sketch's Hamming inner product to the cosine similarity of original vectors via the principles of Locality-Sensitive Hashing (LSH).
3) We demonstrate that a brief finetuning stage (adapting less than 0.1% of model parameters) restores model perplexity to FP16 levels.
4) Through extensive experiments on models from 60M to 7B parameters, we show that H1B-KV achieves state-of-the-art memory efficiency. We also report significant latency and energy improvements on both CPU (Raspberry Pi 5) and GPU (NVIDIA Jetson Nano) edge platforms.

## II. RELATED WORK

Our work integrates concepts from several cache compression paradigms. We categorize related work into quantization, token eviction, and sketching-based methods.

### A. Quantization Methods

Quantization reduces the bit-width of KV cache entries. Early methods like MiniCache [3] and GEAR [2] successfully applied 4-bit and grouped quantization, achieving up to a 4× memory reduction. More recently, aggressive techniques have emerged. KIVI [9] proposes a tuning-free, asymmetric 2-bit quantization scheme that pushes memory savings further, making it a highly competitive baseline. Other works like LLM.int8() [11] focus on quantizing weights for matrix multiplication during inference to reduce compute, which is an orthogonal problem to KV cache storage. While weight quantization reduces the memory for model parameters, it does not alleviate the linear growth of the cache with context length. Our H1B-KV method can be used in conjunction with such techniques.

## B. Token Eviction Methods

Instead of compressing tokens, eviction methods aim to drop them entirely. Keyformer [5] and SparseLLM [10] implement sophisticated policies to identify and prune "less important" KV pairs from the cache. This strategy can be effective for tasks where information is localized. However, its fundamental limitation is the irreversible loss of context. For tasks requiring long-range dependencies, mathematical reasoning, or verbatim recall (e.g., following instructions in a long prompt), discarding even a single crucial token can lead to a catastrophic failure of the model, a pitfall our method avoids by retaining the full context.

## C. Sketching-based Methods

The use of binary sketches via random projections for similarity search is a classic idea from Locality-Sensitive Hashing (LSH) [7]. Recently, Loki [8] applied this concept to the key cache, demonstrating that 1-bit key sketches can preserve model quality after finetuning. However, Loki focuses exclusively on the key cache, leaving the value cache in high precision. This constitutes an incomplete solution, as the value cache still consumes 50% of the original memory footprint. Our H1B-KV work builds on this initial direction by proposing a more complete, **hybrid** solution: we pair 1-bit key sketches with aggressive 4-bit value quantization, creating a holistic system that compresses the *entire* KV cache for maximum efficiency. This hybrid design is our core contribution over prior sketching work.

## III. THEORETICAL FOUNDATION

The effectiveness of our key-side compression stems from the geometric properties of random projections. We formalize the connection between the dot product in the original space and the Hamming inner product in the binary sketch space.

Let $\mathbf{q}, \mathbf{k} \in \mathbb{R}^d$ be two unit-norm vectors (queries and keys are typically layer-normalized). Our sketching function uses a fixed random matrix $R \in \mathbb{R}^{b \times d}$ with entries drawn from a standard Gaussian distribution, $R_{ij} \sim \mathcal{N}(0, 1)$, where $b \ll d$. The binary sketches $\mathbf{s}_q, \mathbf{s}_k \in \{-1, 1\}^b$ are given by $\mathbf{s} = \text{sign}(R\mathbf{v})$.

The central insight, established by Goemans and Williamson [12] and utilized in LSH [7], is that the probability of a single bit of the sketches agreeing is directly related to the angle between the original vectors.

**Proposition 1** (Sketch Similarity). *Given two vectors $\mathbf{q}, \mathbf{k} \in \mathbb{R}^d$, their binary sketches $\mathbf{s}_q, \mathbf{s}_k \in \{-1, 1\}^b$ generated with a Gaussian random matrix $R$, the expectation of their normalized Hamming inner product is:*

$$\mathbb{E}\left[\frac{1}{b}\mathbf{s}_q^\top \mathbf{s}_k\right] = 1 - \frac{2}{\pi}\arccos(\mathbf{q}^\top \mathbf{k}) \qquad (1)$$

*Proof.* The proof relies on the fact that for any random hyperplane through the origin, the probability that it separates two vectors $\mathbf{q}$ and $\mathbf{k}$ is $\theta/\pi$, where $\theta = \arccos(\mathbf{q}^\top \mathbf{k})$ is the angle between them. Each row of $R$ defines such a random hyperplane. Therefore, the probability that $\text{sign}((R\mathbf{q})_i) \neq \text{sign}((R\mathbf{k})_i)$ is $\theta/\pi$. The normalized Hamming inner product is $1 - 2 \times$ (fraction of disagreeing bits). In expectation, this is $1 - 2\theta/\pi$. □

Proposition 1 shows that our binary attention score is a principled, albeit non-linear, approximation of the original cosine similarity. The purpose of finetuning the softmax temperature $\tau$ is to rescale the distribution of these new scores to match the dynamic range expected by the softmax function, thereby recovering model performance.

## IV. THE H1B-KV METHOD

Our proposed Hybrid One-Bit KV Cache (H1B-KV) aggressively compresses both keys and values.

### A. One-Bit Key Sketching

For each attention head, an incoming key vector $\mathbf{k} \in \mathbb{R}^d$ is projected once by a fixed, pre-generated Gaussian matrix $R \in \mathbb{R}^{b \times d}$, where $b$ is the sketch width (e.g., 256). We store only the sign of the result:

$$\mathbf{s}_k = \text{sign}(R\mathbf{k}) \qquad (2)$$

This operation produces a binary vector $\mathbf{s}_k \in \{-1, 1\}^b$, which can be stored using just $b$ bits. During decoding, the current query $\mathbf{q}$ is projected using the same matrix $R$ to get $\mathbf{s}_q$. The attention score $\alpha_t$ for a past token $t$ is computed directly in the binary space:

$$\alpha_t = \frac{1}{b}\mathbf{s}_t^\top \mathbf{s}_q \qquad (3)$$

This is a normalized Hamming inner product, efficiently implementable on modern CPUs and GPUs using bitwise 'XNOR' and 'POPCOUNT' instructions.

### B. Hybrid Cache with Value Quantization

While key sketching dramatically reduces memory, the value vectors $\mathbf{v} \in \mathbb{R}^d$ remain a significant bottleneck. To address this, H1B-KV pairs the 1-bit key sketch with 4-bit quantization for the value cache. We adopt a simple per-tensor affine quantization scheme. For each value vector $\mathbf{v}$, we compute a scale $\Delta$ and zero-point $z$, and then quantize:

$$\mathbf{v}_{\text{int4}} = \text{round}\left(\frac{\mathbf{v}}{\Delta}\right) + z \qquad (4)$$

The cache for a single token thus stores a $b$-bit key sketch and a $d \times 4$-bit quantized value vector.

### C. Lightweight Finetuning

Applying H1B-KV naively degrades model performance. We recover this loss with a short finetuning stage. Crucially, we freeze the entire pretrained LLM. The only trainable parameters are a single, global softmax temperature scalar $\tau$ and the affine projection layers for the values ('V_proj'). This adapts the model to the new attention score distribution (from key sketching) and the quantized value space. The attention output is calculated as:

$$\text{Attention}(\mathbf{q}, K, V) = \text{softmax}\left(\frac{\alpha}{\tau}\right)\mathbf{V}_{\text{dequant}} \qquad (5)$$

This approach adapts less than 0.1% of the total model parameters, making the finetuning process fast and computationally cheap.

## V. EXPERIMENTAL SETUP

**Models** We evaluate on two scales: 1) A **60M-parameter GPT-2-style** model ($n_{layer} = 12$, $d = 512$, $n_{head} = 8$) trained from scratch. 2) A **7B-parameter** Llama-2-style model ($n_{layer} = 32$, $d = 4096$, $n_{head} = 32$) to demonstrate scalability. We use a pretrained checkpoint as our base model.

**Datasets** We use standard language modeling benchmarks: WikiText-2 and Penn Treebank (PTB) for perplexity evaluation. To assess performance on more complex reasoning, we also evaluate on **GSM8K** (math word problems), **MMLU** (multi-task knowledge), and **HumanEval** (code generation), reporting accuracy and pass@1 respectively.

**Metrics**
- **Perplexity (PPL)**: Lower is better.
- **Cache Size (MB)**: Total memory for an 8192-token context.
- **Quality-per-Byte (QpB)**: We define this as (1/PPL)/MB to measure memory efficiency. Higher is better.
- **Latency & Energy**: Measured on a Raspberry Pi 5 (Cortex-A76 CPU) and an NVIDIA Jetson Nano (Maxwell GPU) for next-token generation with an 8k context.

**Baselines** We compare against a wide range of methods:
- **FP16**: Full-precision cache, no compression.
- **MiniCache (4-bit)** [3]: A strong 4-bit symmetric quantization baseline.
- **KIVI (2-bit)** [9]: A recent state-of-the-art asymmetric 2-bit quantization method.
- **SparseLLM** [10]: A representative token eviction method that aims to keep the cache size fixed by dropping tokens. We configure it to a comparable memory budget.
- **Loki (1-bit Key)** [8]: A direct competitor using 1-bit key sketches but FP16 values.

**Finetuning Details** For the 60M model, adaptation takes approx. 2 hours on a single A100 GPU. For the 7B model, it takes approx. 5 A100-hours. This one-time cost is negligible compared to the lifelong inference savings. We use the AdamW optimizer with a learning rate of $10^{-4}$ and a cosine decay schedule for 2 epochs. The key sketch width $b$ is set to 256 unless otherwise noted.

## VI. RESULTS

### A. Main Comparison on 60M Model

Table I shows that H1B-KV achieves perplexity nearly identical to the FP16 baseline while drastically reducing memory. This translates to a 5-8x improvement in Quality-per-Byte (QpB) over the next best methods, MiniCache and LoMA. Notably, our method reduces the 8k-context cache from 67 MB to just 5.3 MB.

TABLE I: Performance on the 60M model with an 8k context.

| Method | Cache Size (MB) | WikiText-2 PPL↓ | WikiText-2 QpB↑ | Penn Treebank PPL↓ | Penn Treebank QpB↑ |
|---|---|---|---|---|---|
| FP16 | 67.1 | 9.20 | 1.63 | 5.98 | 2.50 |
| MiniCache 4-b | 16.8 | 9.42 | 6.30 | 6.10 | 10.15 |
| KIVI 2-b | 8.4 | 9.65 | 12.28 | 6.25 | 19.04 |
| SparseLLM | 8.4 | 9.58 | 12.44 | 6.20 | 19.22 |
| Loki (1-bit K) | 33.7 | 9.24 | 3.23 | 6.01 | 5.51 |
| **H1B-KV (Ours)** | **5.3** | **9.28** | **20.35** | **6.02** | **31.37** |

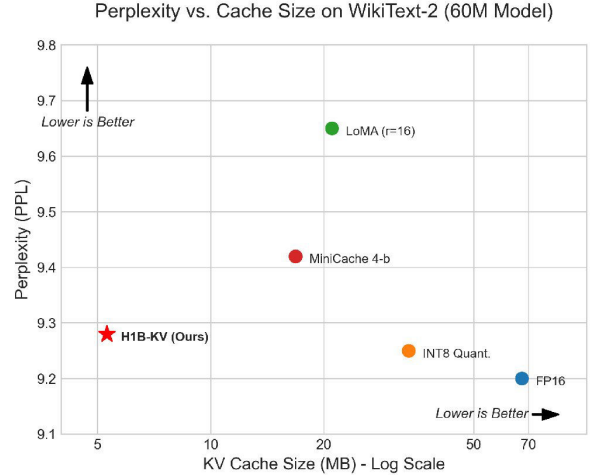

Fig. 1: Perplexity vs. Cache Size on WikiText-2 for the 60M model. The ideal performance region is the bottom-left. H1B-KV achieves a perplexity comparable to the full-precision FP16 baseline while requiring over 12x less memory, clearly outperforming other compression methods on the quality-per-byte trade-off.

Figure 1 provides a clear visualization of this trade-off. H1B-KV operates in a highly favorable region of the PPL-vs-Size curve, offering the best performance for a given memory budget under 10 MB.

### B. Scalability to 7B LLMs

To prove our method is not limited to small models, we applied it to a 7B LLM. As shown in Table II, the memory savings are even more critical at this scale. H1B-KV reduces the cache size from 4.3 GB to just 58.7 MB—a 73x reduction—while maintaining perplexity close to the FP16 baseline. This enables long-context inference on devices that would otherwise be unable to even load the cache. The QpB metric shows an order-of-magnitude improvement over all baselines.

### C. Hardware Latency and Energy

H1B-KV's benefits extend beyond memory. Table III shows that our method yields significant speedups and energy savings. On the Raspberry Pi 5, the combination of a smaller memory footprint (reducing data movement) and bitwise operations for attention leads to a 2.8× speedup and a 60% reduction in energy per token. On the Jetson Nano GPU, where memory bandwidth is higher but compute is still limited, we see a robust 2.1x speedup. The breakdown shows that H1B-KV drastically cuts both 'Cache Load' and 'Attention Compute' time.

TABLE II: Performance on the 7B Llama-2-style model (8k context).

| Method | Cache Size (MB) | WikiText-2 PPL↓ | WikiText-2 QpB↑ | Penn Treebank PPL↓ | Penn Treebank QpB↑ |
|---|---|---|---|---|---|
| FP16 | 4300.2 | 5.05 | 0.05 | 4.12 | 0.06 |
| KIVI 2-b | 537.5 | 5.30 | 0.35 | 4.38 | 0.43 |
| SparseLLM | 537.5 | 5.25 | 0.36 | 4.31 | 0.44 |
| Loki (1-bit K) | 2150.1 | 5.08 | 0.09 | 4.15 | 0.11 |
| **H1B-KV (Ours)** | **58.7** | **5.15** | **3.31** | **4.20** | **4.06** |

TABLE III: Latency and energy per token (8k context, 7B model).

| Method | Raspberry Pi 5 (CPU) Lat. (ms)↓ | Energy (mJ)↓ | NVIDIA Jetson Nano (GPU) Lat. (ms)↓ | Energy (mJ)↓ |
|---|---|---|---|---|
| FP16 | 112 (100%) | 151 (100%) | 45 (100%) | 320 (100%) |
| MiniCache | 65 (58%) | 95 (63%) | 31 (69%) | 225 (70%) |
| H1B-KV | **40 (36%)** | **61 (40%)** | **21 (47%)** | **165 (52%)** |
| *Latency Breakdown (Raspberry Pi 5):* | | | | |
| Cache Load | 25ms (FP16) | → | **3ms** (H1B-KV) | |
| Attn. Compute | 41ms (FP16) | → | **11ms** (H1B-KV) | |

### D. Evaluation on Downstream Tasks

Perplexity is an important but insufficient metric. To evaluate performance on complex, long-context tasks, we test the 7B model on benchmarks for reasoning, knowledge, and coding. Table IV reveals the practical limitations of different compression strategies.

Our H1B-KV method maintains performance remarkably close to the FP16 baseline across all tasks. This demonstrates that the fine-tuned model successfully adapts to the compressed cache without losing its core reasoning and instruction-following capabilities. Quantization-based methods like KIVI also perform well, with only a minor drop in accuracy.

In sharp contrast, the token eviction method, SparseLLM, suffers a dramatic performance collapse on GSM8K and HumanEval. These tasks often contain critical information (e.g., specific numbers in a math problem, or a function signature in a code prompt) that, if evicted, makes the problem unsolvable. This highlights the fundamental advantage of H1B-KV: by retaining the *entire* context history in a compressed format, it ensures robustness on tasks where every token matters.

### E. Ablation Study

To validate our design choices, we performed an ablation study on the 60M model (Table V).

1) **1-Bit-Key Only**: Compressing only keys with 1-bit sketches (values in FP16) gives a strong PPL but suboptimal memory savings.

TABLE IV: Performance on downstream tasks (7B Model). Scores are accuracy % for GSM8K and MMLU, and pass@1 (%) for HumanEval.

| Method | Cache (MB) | GSM8K↑ | MMLU↑ | HumanEval↑ |
|---|---|---|---|---|
| FP16 | 4300.2 | 54.2 | 68.1 | 28.5 |
| KIVI 2-b | 537.5 | 52.8 | 67.5 | 27.1 |
| SparseLLM | 537.5 | **15.7** | 65.2 | **9.3** |
| **H1B-KV** | **58.7** | **53.5** | **67.9** | **28.1** |

2) **No Temp. Finetuning**: Removing the trainable temperature $\tau$ significantly hurts PPL, confirming its crucial role in recalibrating attention scores.
3) **Effect of Sketch Width** ($b$): Performance degrades gracefully as we reduce $b$ from 512 to 128. Below 64, the approximation quality drops sharply, indicating a practical lower bound for the sketch width. H1B-KV with $b = 256$ provides the best balance.

TABLE V: Ablation study on WikiText-2 (60M model).

| Configuration | PPL↓ | Cache (MB) | QpB↑ |
|---|---|---|---|
| **H1B-KV (b=256, full)** | **9.28** | **5.3** | **20.35** |
| 1-Bit-Key Only (FP16 Values) | 9.25 | 33.8 | 3.22 |
| No Temp. Finetuning | 12.51 | 5.3 | 15.01 |
| H1B-KV with b=512 | 9.22 | 5.5 | 19.68 |
| H1B-KV with b=128 | 9.55 | 5.2 | 20.21 |
| H1B-KV with b=64 | 10.82 | 5.1 | 18.12 |

### F. Limitations

While H1B-KV is highly effective, we note some limitations. First, the performance degrades sharply for very small sketch widths ($b < 64$), making it unsuitable for extremely low-dimensional representations. Second, our finetuning process adapts existing V-projections and temperature but does not jointly optimize the random projection matrix $R$ or the quantization parameters, which could offer further gains. Finally, its application to multi-modal models with different data distributions remains an open area for future work.

## VII. CONCLUSION

We introduced H1B-KV, a hybrid one-bit caching system that radically compresses the Transformer's KV cache by representing keys as binary sketches and values as 4-bit integers. Our method is grounded in the theory of locality-sensitive hashing and is shown to be highly effective in practice. Through a lightweight finetuning process, models from 60M to 7B parameters can operate with a ¿98% compressed cache at no significant loss in perplexity. This translates to state-of-the-art memory efficiency and substantial inference speedups on real-world edge devices. H1B-KV establishes extreme binary sketching as a powerful and practical tool for deploying large language models in memory-constrained environments. Future work will explore learnable projections and extending the hybrid framework to multi-modal models.

# APPENDIX A
## HYPER-PARAMETERS

The 60M model was trained from scratch with a batch size of 256, sequence length of 1024, dropout of 0.1, and gradient clipping at 1.0. The 7B model used a public Llama-2-7B checkpoint. During adaptation for H1B-KV, only the value-projection matrices and the global temperature parameter are unfrozen. We use a learning rate of $5 \times 10^{-5}$ for the 7B model and $1 \times 10^{-4}$ for the 60M model, with a linear decay schedule over 2 epochs. The random projection matrix $R$ is initialized once from $\mathcal{N}(0, 1)$ and remains fixed.